\documentclass[conference]{IEEEtran}
\IEEEoverridecommandlockouts
\usepackage{cite}
\usepackage{amsmath,amssymb,amsfonts}
\usepackage{algorithmic}
\usepackage{graphicx}
\usepackage{subcaption}
\usepackage{booktabs}
\usepackage{textcomp}
\usepackage{xcolor}
\usepackage{multirow} 
\usepackage{array} 

\def\BibTeX{{\rm B\kern-.05em{\sc i\kern-.025em b}\kern-.08em
    T\kern-.1667em\lower.7ex\hbox{E}\kern-.125emX}}
    
\begin{document}


\title{Multi-Robot Coordination in V2X Environments\\
}

\author{
  \IEEEauthorblockN{
    John Pravin Arockiasamy\IEEEauthorrefmark{1} and
    Alexey Vinel\IEEEauthorrefmark{1}\IEEEauthorrefmark{2}
  }

  \IEEEauthorblockA{
    \IEEEauthorrefmark{1}Karlsruhe Institute of Technology, 
    Karlsruhe, Germany \\
    Email: alexey.vinel@kit.edu
  }

  \IEEEauthorblockA{
    \IEEEauthorrefmark{2}Halmstad University, Halmstad, Sweden
    \\
    Email: alexey.vinel@hh.se
  }
}

\maketitle
\begingroup
\renewcommand\thefootnote{}
\footnotetext{This paper has been accepted for publication at the IEEE Intelligent Transportation Systems Conference (ITSC) 2026.}
\endgroup

\begin{abstract}
This paper presents a Vehicle-to-Everything (V2X) communication framework that enables decentralized cooperation among social robots operating in complex urban traffic environments. Building on ETSI Cooperative Awareness and Maneuver Coordination services, the framework introduces two robot-centric facility-layer services: the \emph{Robot Awareness Service} (RAS) and the \emph{Robot Maneuver Coordination Service} (RMCS), realized through the Robot Awareness Message (RAM) and the Robot Maneuver Coordination Message (RMCM), respectively. RAS enables role-aware, task-oriented robot awareness while integrating externally detected Vulnerable Road Users (VRUs), including non-V2X pedestrians, into cooperative awareness. RMCS supports event-driven, low-latency coordination of robot maneuvers under explicitly established roles, without centralized infrastructure or prior pairing. A real-world proof of concept demonstrates deterministic multi-robot coordination between a humanoid robot and a quadrupedal robot assisting a pedestrian during a road-crossing scenario, governed by a formally specified finite-state coordination model. Complementary simulations evaluate robot-mediated VRU clustering in mixed V2X environments, showing that RAS-based clustering integrates non-V2X VRUs in safety-critical areas while reducing redundant transmissions from V2X-enabled VRUs, thereby lowering channel load. Together, the proposed services provide a scalable and standards-aligned foundation for integrating cooperative robots into future Connected, Cooperative, and Automated Mobility ecosystems.
\end{abstract}

\begin{IEEEkeywords}
V2X, cooperative robots, maneuver coordination, vulnerable road users, connected mobility, pedestrian safety.
\end{IEEEkeywords}

\section{Introduction}

Robotic technologies have matured rapidly across domains such as manufacturing, healthcare, and domestic assistance. However, their integration into outdoor urban traffic systems is still developing and remains underexplored. In future smart cities, social robots could support public assistance tasks including pedestrian guidance, traffic mediation, construction zone management, and temporary safety enforcement. Operating in road traffic environments poses unique challenges, as robots must safely coexist and cooperate with vehicles, pedestrians, and infrastructure in highly dynamic and safety-critical conditions~\cite{yu2024understanding}.

In parallel, the gradual deployment of automated and connected vehicles is reshaping urban mobility. While automation promises significant safety benefits, the transition toward fully automated traffic systems will be prolonged and characterized by heterogeneous participants, including human-driven vehicles and unprotected Vulnerable Road Users (VRUs)~\cite{Milford2024}. Achieving long-term road safety objectives such as Vision Zero therefore requires adaptive and cooperative frameworks capable of harmonizing interactions between autonomous agents and human actors within Connected, Cooperative, and Automated Mobility (CCAM) ecosystems. Existing traffic regulation concepts, however, are typically designed either for human-driven or fully automated traffic, with limited consideration for their coexistence in mixed environments~\cite{Bied2024}.

More work has explored humanoid robots acting as traffic controllers, often using gestures-based or visual interaction~\cite{zhao2019,ghaffarControllingTrafficHumanoid, ben2024mika}. The IPA2X project\footnote{https://ipa2x.eu/} introduced a mobile robotic traffic light to improve pedestrian safety at crossings, and subsequent proof of concept (POC) deployments by our research team have demonstrated that robots can interact effectively with pedestrians and vehicles in real-world settings~\cite{arockiasamy2026robopolsocialroboticsmeets,IAVVC2025}. However, these studies also reveal a key limitation: a single robot can address only a narrow subset of traffic situations, whereas complex scenarios such as two-way crossings require coordinated teams of robots, similar to how human traffic stewards collaborate in practice.

Despite this need, no standardized V2X-based mechanism currently enables robots to coordinate roles, intentions, and maneuvers within existing CCAM systems. ETSI facility-layer messages--including Cooperative Awareness Message (CAM)~\cite{ETSI_V2_CAM}, Vulnerable road users Awareness Message (VAM)~\cite{etsi2023ts103300-3}, Collective Perception Message (CPM)~\cite{ETSI_NEW_CPM}, and Maneuver Coordination Message  (MCM)~\cite{etsi_tr103578_v0211_2024}--are designed primarily for vehicles and VRUs. They do not capture robot-specific operational roles, task intent, or the integration of externally sensed, non-V2X VRUs, and provide limited support for role-based coordination among heterogeneous robotic agents.

This paper addresses this gap by introducing a robot-centric Vehicle-to-Everything (V2X) communication framework that integrates robots as first-class participants in vehicular ad hoc networks (VANETs). V2X-enabled robots can already be deployed, as multi-robot communication does not rely on full V2X adoption by other road users. Building on ETSI facility-layer principles, we extend Cooperative Awareness and Maneuver Coordination through two robot-specific services: the \emph{Robot Awareness Service} (RAS) and the \emph{Robot Maneuver Coordination Service} (RMCS). RAS provides role-aware, and task-oriented robot awareness while integrating the externally sensed VRUs into the cooperative domain. RMCS supports event-driven, low-latency coordination of robot maneuvers. Both services operate fully decentralized manner via dedicated messages--the Robot Awareness Message (RAM) and the Robot Maneuver Coordination Message (RMCM)--without centralized control, prior pairing, or infrastructure dependency.

The proposed framework is validated through a real-world POC in which a humanoid robot and a quadrupedal robot coordinate to assist a pedestrian during a road-crossing scenario, following a formally defined finite-state coordination model and relying only on RAM and RMCM exchanges. In addition, simulation-based evaluations assess the scalability of robot-mediated VRU integration, focusing on VRU clustering and its impact on channel utilization in mixed V2X environments.

The main contributions of this paper are:
\begin{itemize}
\item The design of robot-centric V2X facility-layer services that extend ETSI standards to support decentralized robot awareness, coordination, and VRU integration.
\item A real-world demonstration of deterministic multi-robot cooperation for pedestrian assistance using heterogeneous robots and V2X communication.
\item A simulation-based evaluation showing that robot-mediated VRU clustering improves cooperative awareness while reducing communication load in VANETs.
\end{itemize}

The remainder of the paper is organized as follows. Section~II introduces the proposed RAS and RMCS and their design rationale. Section~III presents the real-world deployment and simulation-based evaluation. Section~IV concludes the paper and outlines directions for future research.

\section{Motivation and Design Rationale for Robot V2X Services}
\label{sec:motivation}

The increasing presence of mobile robots in urban environments introduces new challenges for CCAM systems. Unlike conventional vehicles or infrastructures, robots operate on a pedestrian scale, dynamically change roles, and frequently interact with VRUs. This section motivates the need for robot-specific V2X services by outlining representative use cases, deriving system requirements, analyzing the limitations of existing ETSI messages, and introducing the proposed messages.

\subsection{Use Cases}
\label{sec:use_cases}

Urban traffic environments increasingly require temporary, adaptive mediation to ensure pedestrian safety, particularly in scenarios such as school crossings. Today, these situations are commonly handled by human traffic stewards who physically intervene in traffic flows, thereby exposing themselves to significant risk in mixed traffic conditions. This work considers the deployment of mobile robots as active traffic mediators capable of jointly assuming these safety-critical roles. Due to their mobility, re-deployability, and social interaction capabilities, robots can be dynamically, positioned at school zones. However, as demonstrated in~\cite{IAVVC2025}, complex traffic situations cannot be reliably managed by a single robot alone, inherently  necessitates explicit multi-robot coordination to distribute roles, synchronize actions, and ensure consistent behavior under dynamic conditions. 

Unlike conventional multi-robot systems that rely on centralized short-range communication (e.g., WLAN or Bluetooth) within structured environments such as warehouses or industrial facilities~\cite{gielis2022critical}, traffic mediation requires robots to operate directly within open urban road systems as cooperative traffic participants. In these safety-critical environments, robots must interact with vehicles, infrastructure, and VRUs under highly dynamic and decentralized conditions, while continuously exchanging awareness information which can be achieved through V2X technology. 

A further challenge arises from the fact that many VRUs do not carry V2X-enabled devices and therefore remain invisible to cooperative awareness systems. Mobile robots equipped with onboard perception can locally detect such participants and act as intermediaries between the physical environment and the V2X domain, enabling their representation within cooperative traffic systems. Beyond
sensing, robots can intervene directly through visual cues,
audible warnings, or gestures to stop a pedestrian before any dangerous situation~\cite{arockiasamy2026robopolsocialroboticsmeets}.
Because the
robot's sensing range moves with it, coverage naturally
extends to regions that fixed sensors cannot observe. Roadside
infrastructure can partially address these challenges, but its
field of view is bounded by the mounting location and provides
no awareness outside that area~\cite{arnold2020cooperative},
which limits its usefulness in temporary or evolving traffic
situations.

Taken together, these characteristics fundamentally distinguish urban traffic mediation from traditional multi-robot applications and directly motivate the functional, performance, and safety requirements formalized in the following section.




\subsection{System Requirements}
\label{sec:requirements}

The identified use cases necessitate the integration of robots into CCAM systems, from which the following requirements are derived:


\begin{itemize}
    \item \textbf{R1: Decentralized Robot Awareness} -- Robots shall transmit awareness messages that reflect their role, task execution state, and sensed VRU context as traffic participants without relying on centralized infrastructure.

    \item \textbf{R2: Multi-Robot Coordination} -- The framework shall support explicit coordination among multiple robots, including role establishment and task execution.
    
    \item \textbf{R3: Support for Non-V2X VRUs} -- Robots shall be able to represent VRUs that do not transmit V2X messages using onboard perception.
    
    \item \textbf{R4: Low-Latency Coordination and Execution} -- Multi-robot coordination mechanisms shall enable timely maneuver execution in safety-critical scenarios.
    
    \item \textbf{R5: Fail-Safe Operation} -- The system shall provide mechanisms to ensure safe behavior under communication loss or execution failure.

    \item \textbf{R6: Backward ETSI Compatibility} -- The solution shall align with existing ETSI ITS-G5 standards to enable seamless integration with current CCAM deployments.
        
    \item \textbf{R7: Security and Privacy Preservation} -- Robot-generated V2X messages shall comply with ETSI ITS-Sec mechanisms.
    
\end{itemize}

\subsection{Why CAM, VAM, CPM, and MCM Are Insufficient}
\label{sec:limitations}

ETSI ITS facility-layer services such as CAM, VAM, CPM, and MCM provide essential awareness and coordination primitives for vehicles and VRUs. However, when applied to robot-mediated traffic scenarios derived from the use cases in Section~\ref{sec:use_cases} and the requirements in Section~\ref{sec:requirements}, these services reveal fundamental limitations.

\begin{itemize}
    \item \textbf{CAM:}  
    CAM conveys periodic state information of traffic participants but lacks semantics for robot-specific characteristics such as robot kinematics, dynamic role assignment, and task execution state. Robots cannot be distinguished from generic stations, nor can they express whether they are requesting, offering, or executing a traffic mediation task.

    \item \textbf{VAM:}  
    VAM represents VRUs that actively transmit their own status. This inherently assumes that pedestrians and other VRUs are V2X-equipped, preventing the inclusion of non-equipped VRUs detected through onboard robot perception, which is essential in mixed traffic.

    \item \textbf{CPM:}  
    CPM enables the exchange of perceived object information and can partially represent clustered VRU awareness~\cite{ETSI_NEW_CPM}. However, in mobile traffic mediation scenarios, robots dynamically approach and manage dense groups of VRUs. Relying solely on object-level CPM transmissions leads to redundant messaging and increased channel load~\cite{chtourou2021context}.

    \item \textbf{MCM:}  
    MCM supports explicit maneuver coordination but is vehicle-centric and tailored to driving-related actions, limiting its applicability to heterogeneous robot coordination tasks. 
\end{itemize}
Existing ETSI facility-layer services therefore address awareness and coordination at the vehicle or object level, but lack the semantic constructs required for robot-centric traffic mediation. While CAM, VAM, and CPM are extensible by design, re-purposing them to encode task-level robot semantics would overload awareness-centric constructs with coordination intent, thereby exceeding their original design scope and reducing interoperability. This gap motivates the introduction of robot-specific facility-layer services that extend ETSI concepts while remaining compatible with the ITS-G5 architecture.




\subsection{Robot Awareness and Maneuver Coordination Service}
\label{sec:ram_rmcm}




To fulfill the identified system requirements, this work introduces two robot-centric facility-layer services: RAS and RMCS. Through these services, the robots are treated as full ITS stations, allowing direct interpretation of existing message sets, such as CAM, VAM, and CPM, for seamless interoperability with existing ETSI ITS systems. Concretely,
each robot receives these messages from surrounding
stations and, in turn, announces its presence via RAM so that vehicles, VRUs, and
infrastructure perceive it through their existing ETSI
systems.

Both services operate fully within the ETSI ITS-G5 security framework and inherit standard ITS-Sec mechanisms, including authentication, integrity protection, and pseudonymized station identifiers. 


\subsubsection{Robot Awareness Message (RAM)}
\label{sec:ram}

 RAM is the service-specific Protocol Data Unit (PDU) realizing the RAS and follows the periodic and event-driven transmission principles as ETSI CAM. Transmission is triggered by robot dynamic state changes, role establishment, or task execution, thereby preserving established channel behavior while extending cooperative awareness to robot-specific semantics. This design maintains full compatibility with the ETSI V2X framework while introducing robot-centric abstractions.

\begin{figure}[t]
\centering
\includegraphics[width=0.7\columnwidth]{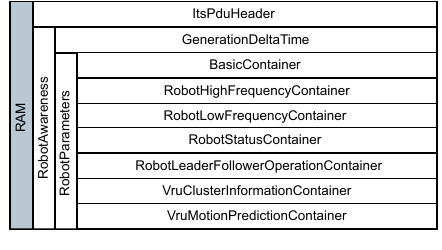}
\caption{Structure of the RAM framework.}
\label{fig:ram_structure}
\end{figure}

\begin{figure}[t]
\centering
\includegraphics[width=0.65\columnwidth]{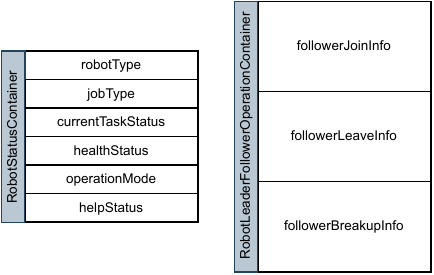}
\caption{Robot status container (left) and Robot coordination containers (right) in the RAM framework.}
\label{fig:robot_status_structure}
\end{figure}

RAM follows a container-based structure conceptually aligned with ETSI CAM~\cite{ETSI_V2_CAM} (Fig.~\ref{fig:ram_structure}). It includes a standard \texttt{ItsPduHeader} and \texttt{GenerationDeltaTime}, a mandatory \texttt{BasicContainer} for station identification and reference position, a mandatory \texttt{RobotHighFrequencyContainer} reporting rapidly changing kinematic parameters, and an optional \texttt{RobotLowFrequencyContainer} carrying slowly varying information.

A key enhancement introduced by RAM is the mandatory \texttt{RobotStatusContainer}, which explicitly captures the robot’s operational context and role within the traffic environment. As illustrated in Fig.~\ref{fig:robot_status_structure}, this container abstracts robot characteristics such as robot category, operational job role, task progression, battery status, and control mode, enabling other ITS entities to reason about robot intent and capability without requiring implementation-specific knowledge.

In addition to robot status dissemination, RAM introduces specialized containers, namely \texttt{RobotLeaderFollowerOperationContainer} for \textit{multi-robot coordination}, and the \texttt{VruClusterInformationContainer} and \texttt{VruMotionPredictionContainer} for \textit{VRU clustering}.

\paragraph{Multi-Robot Coordination Process}
\label{sec:RCP}

RAM enables explicit multi-robot coordination through a lightweight leader-follower mechanism supported by dedicated coordination containers (Fig.~\ref{fig:robot_status_structure}). Coordination is initiated when a robot signals the need for assistance via the \texttt{helpStatus} field within \texttt{RobotStatusContainer}. Nearby robots capable of supporting the task respond to join via \texttt{RobotLeaderFollowerOperationContainer}, leading to the formation of a temporary coordination group.


The requesting robot takes the leader role, while assisting robots join as followers. This mechanism supports coordination of multiple follower robots under a single leader. Even after joining, the follower robots continue to transmit independent RAM, preserving their situational awareness to other road users throughout the coordination process. The leader uses this RAM to maintain an abstracted view of the followers and to support efficient group management.

Task execution and maneuver-level coordination associated with the assigned job are conducted through RMCM, as described in Section~\ref{sec:RMCM}. Coordination is explicitly terminated either upon task completion or via release signaling.

\paragraph{VRU Clustering Process}

RAM further ensures VRU safety by supporting VRU clustering through the \texttt{VruClusterInformationContainer} and \texttt{VruMotionPredictionContainer}, which adhere to the ETSI VAM~\cite{etsi2023ts103300-3}. These containers allow robots to maintain aggregated representations of nearby VRUs and disseminate their collective state within a single message.


Unlike ETSI VAM, which relies on explicit cluster negotiation among VRUs via \texttt{VruClusterOperationContainer}~\cite{etsi2023ts103300-3}, RAM adopts an implicit clustering strategy~\cite{valle2025non}. Robots automatically act as cluster heads and aggregate detected VRUs within a bounded onboard perception radius, without requiring participation from the VRUs themselves. VRU representation is limited to locally observed state and excludes persistent identifiers or extended trajectories. 

This approach provides two key advantages: (1) non-V2X-enabled pedestrians can be integrated into the V2X domain via onboard perception sensors integrated into the robotic platform, and (2) V2X-enabled VRUs benefit from reduced redundant transmissions, as clustering suppresses individual VAM broadcasts~\cite{etsi2023ts103300-3}  in favor of a unified RAM. VAM suppression is applied conservatively and locally to VRUs physically represented by a robot, without modifying VAM semantics or requiring changes to VRU devices. 

\subsubsection{Robot Maneuver Coordination Message (RMCM)}
\label{sec:RMCM}

\begin{figure}[t]
\centering
\includegraphics[width=0.7\columnwidth]{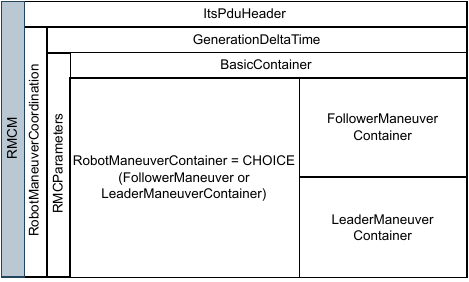}
\caption{Structure of the RMCM framework.}
\label{fig:rmcm_structure}
\end{figure}

RMCM, as shown in Fig.~\ref{fig:rmcm_structure}, is the service-specific PDU used to realize the RMCS, enabling event-driven bidirectional exchange of maneuver instructions and execution feedback among coordinated robots once coordination relationships have been established via RAM. 

\begin{figure}[t]
\centering
\includegraphics[width=0.7\columnwidth]{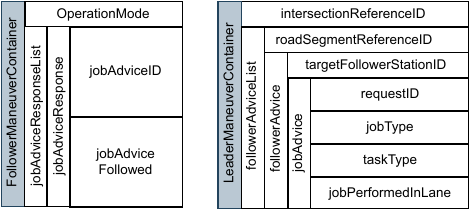}
\caption{Leader and follower maneuver containers in the RMCM framework.}
\label{fig:rmcm_master_slave_structure}
\end{figure}

RMCM distinguishes between leader and follower roles using \texttt{LeaderManeuverContainer} and \texttt{FollowerManeuverContainer}, respectively (Fig.~\ref{fig:rmcm_master_slave_structure}). The leader broadcasts task-specific maneuver instructions with \texttt{requestID} contextualized by road segment or intersection identifiers, while followers respond with execution status and confirmation  with \texttt{jobAdviceID} (derived from the leader's \texttt{requestID}). This separation allows a leader robot to orchestrate multiple followers using
\texttt{targetFollowerStationID}, while ensuring that maneuver commands remain
context-aware and aligned with individual follower capabilities. RMCM generalizes infrastructure-assisted coordination~\cite{correa2019infrastructure} concepts to support decentralized, robot-driven collaboration in dynamic urban environments.



\section{Scenario Illustration and Results}
\label{sec:scenarios}

This section illustrates the practical realization of the proposed RAS and RMCS through representative scenarios derived directly from the use cases in Section~\ref{sec:use_cases}. The objective is to demonstrate decentralized, role-aware multi-robot coordination for pedestrian safety and evaluates the impact of robot-mediated VRU integration and clustering on scalability and communication efficiency in mixed V2X environments.

The first scenario provides a POC for multi-robot coordination in a school-crossing-like setting, where two robots act as active traffic mediators traditionally performed by human stewards. The second scenario evaluates VRU safety in mixed V2X environments, focusing on the effectiveness of robot-mediated VRU clustering using mobile sensing and communication.

\subsection{Multi-Robot Coordination for Pedestrian Assistance}
\label{sec:multi_robot_coordination}

This scenario instantiates the multi-robot traffic mediation use case described in Section~\ref{sec:use_cases}, in which coordinated robots jointly manage pedestrian crossings in safety-critical urban environments. Reflecting typical school-crossing operations, multiple robots cooperate to regulate vehicle flow, guide pedestrians, and maintain consistent situational awareness under dynamic traffic conditions.


As part of the proposed POC, an experimental study is conducted to evaluate coordinated operation between a humanoid robot (ARI, PAL Robotics\footnote{https://pal-robotics.com/robot/ari/}) and a quadrupedal robot (RoboDog-Go2, Unitree Robotics\footnote{https://shop.unitree.com/en-de/products/unitree-go2}) deployed in a controlled environment without live vehicular traffic to ensure safety and repeatability. Both robots are equipped with IEEE~802.11p-based dedicated short-range communication V2X on-board unit from Herman\footnote{https://www.herman.cz/en/} and support the proposed RAS and RMCS. The heterogeneous robot pair reflects realistic deployments where robots with complementary mobility and sensing capabilities cooperate to fulfill a shared safety task.

\begin{figure}[t]
\centering
\includegraphics[width=0.9\columnwidth]{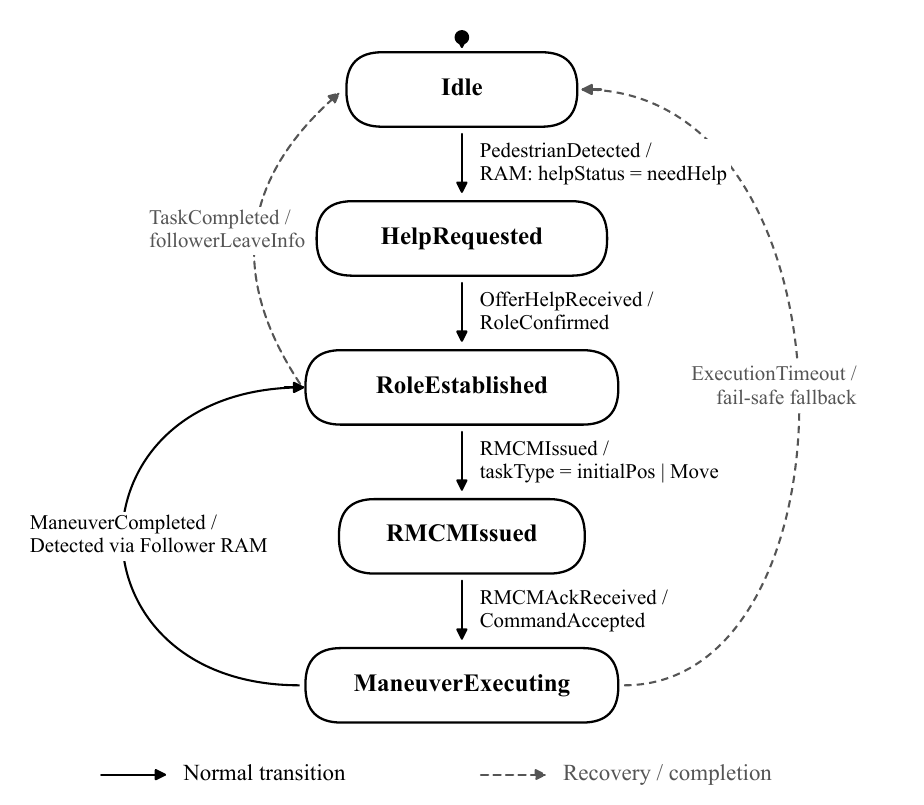}
\caption{FSM governing decentralized multi-robot coordination using RAM and RMCM.}
\label{fig:FSM}
\end{figure}

The coordination protocol is realized as a distributed, event-driven finite-state machine (FSM) executed independently by each robot~(Fig.~\ref{fig:FSM}). Explicit acknowledgment, execution, and completion states ensure deterministic behavior, bounded coordination latency, and fail-safe fallback under communication loss or execution timeouts. Coordination emerges exclusively through RAM and RMCM exchanges directly reflecting the system requirements defined in Section~\ref{sec:requirements}.

\begin{figure*}[t]
    \centering
    \setlength{\tabcolsep}{1pt}
    \renewcommand{\thesubfigure}{\alph{subfigure}}

    \subfloat[Stage 1: Initialization]{%
        \includegraphics[width=0.19\textwidth]{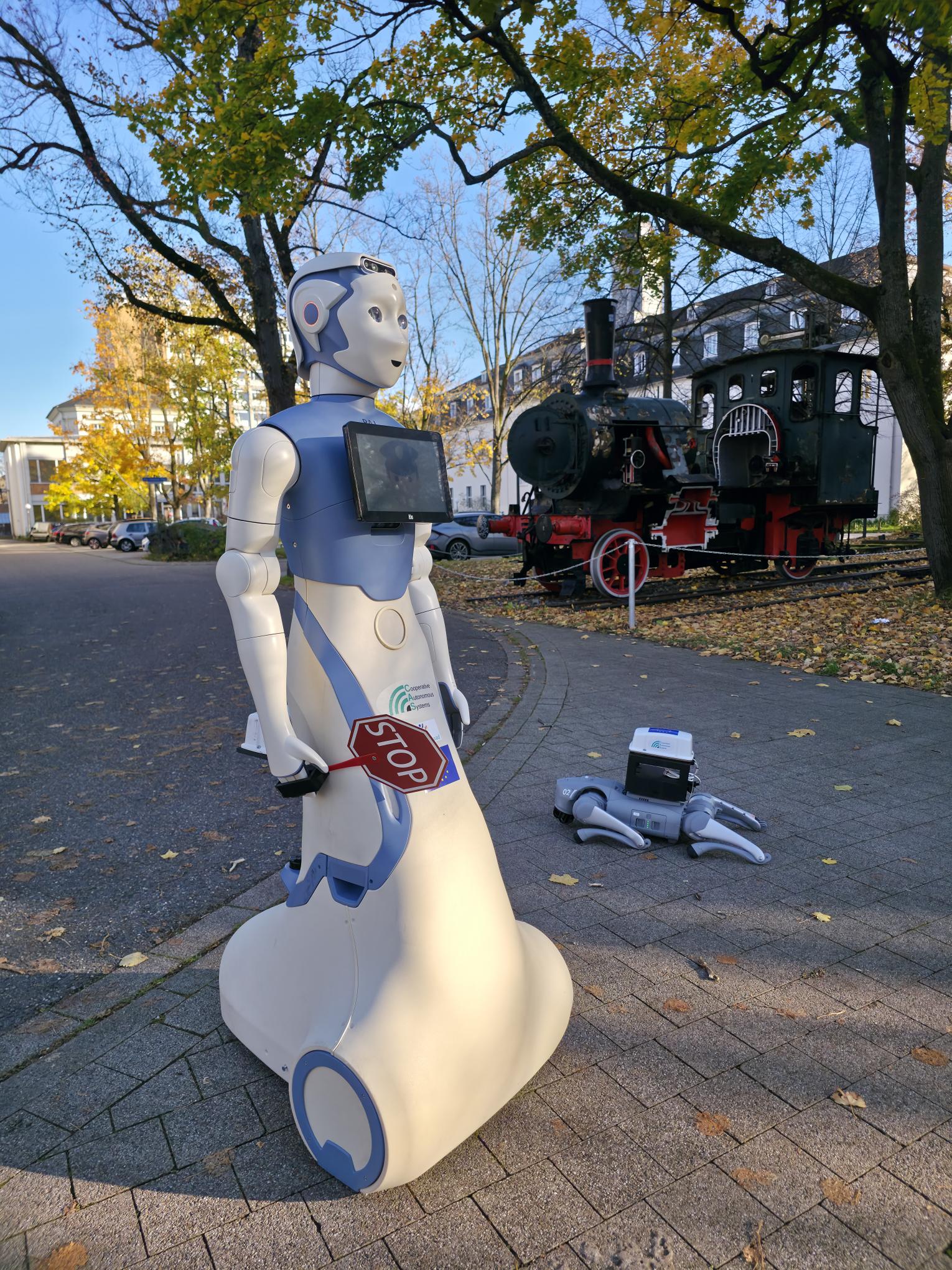}}
    \hfill
    \subfloat[Stage 2: Help Establishment]{%
        \includegraphics[width=0.19\textwidth]{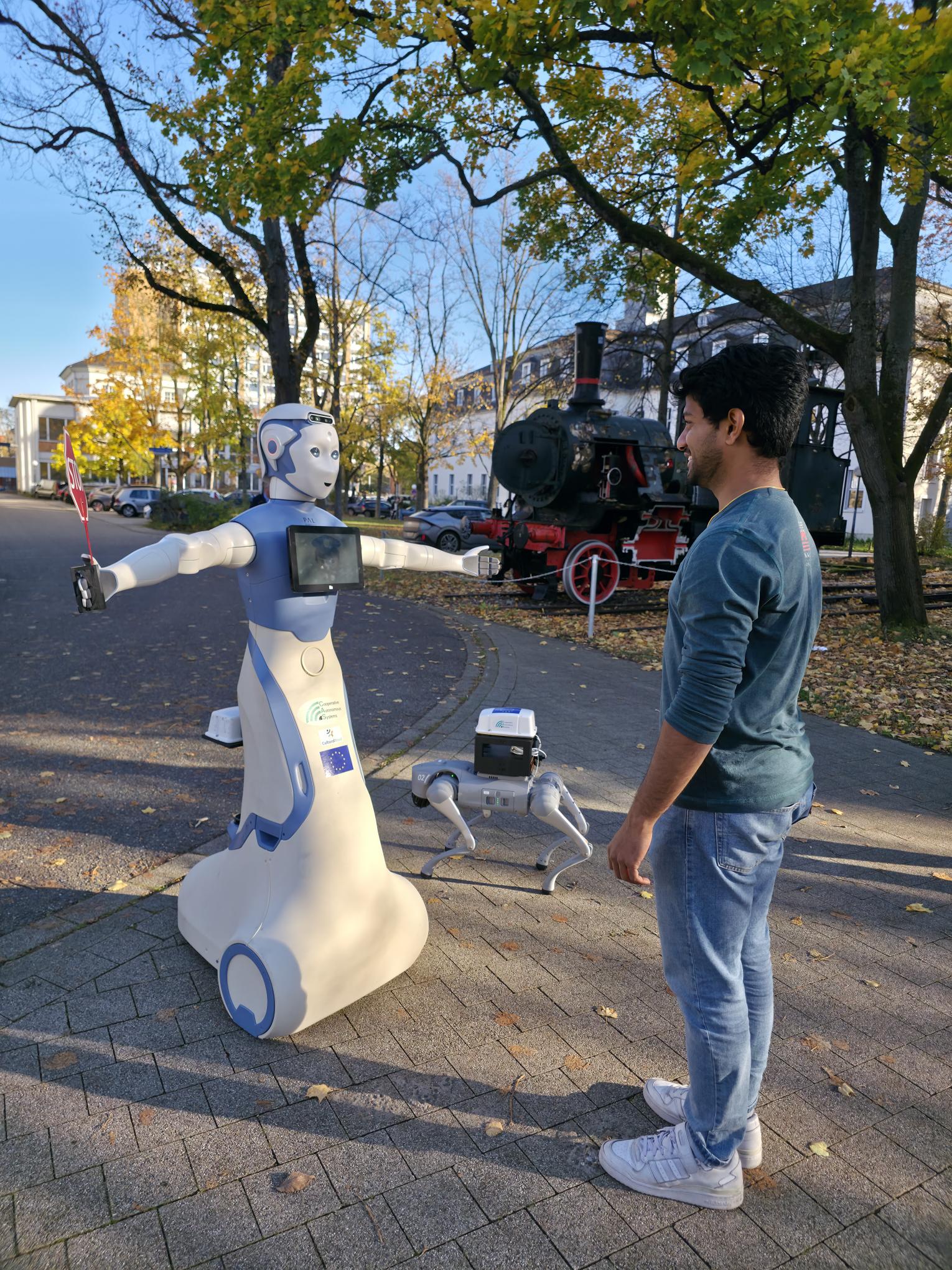}}
    \hfill
    \subfloat[Stage 3: Role Assignment]{%
        \includegraphics[width=0.19\textwidth]{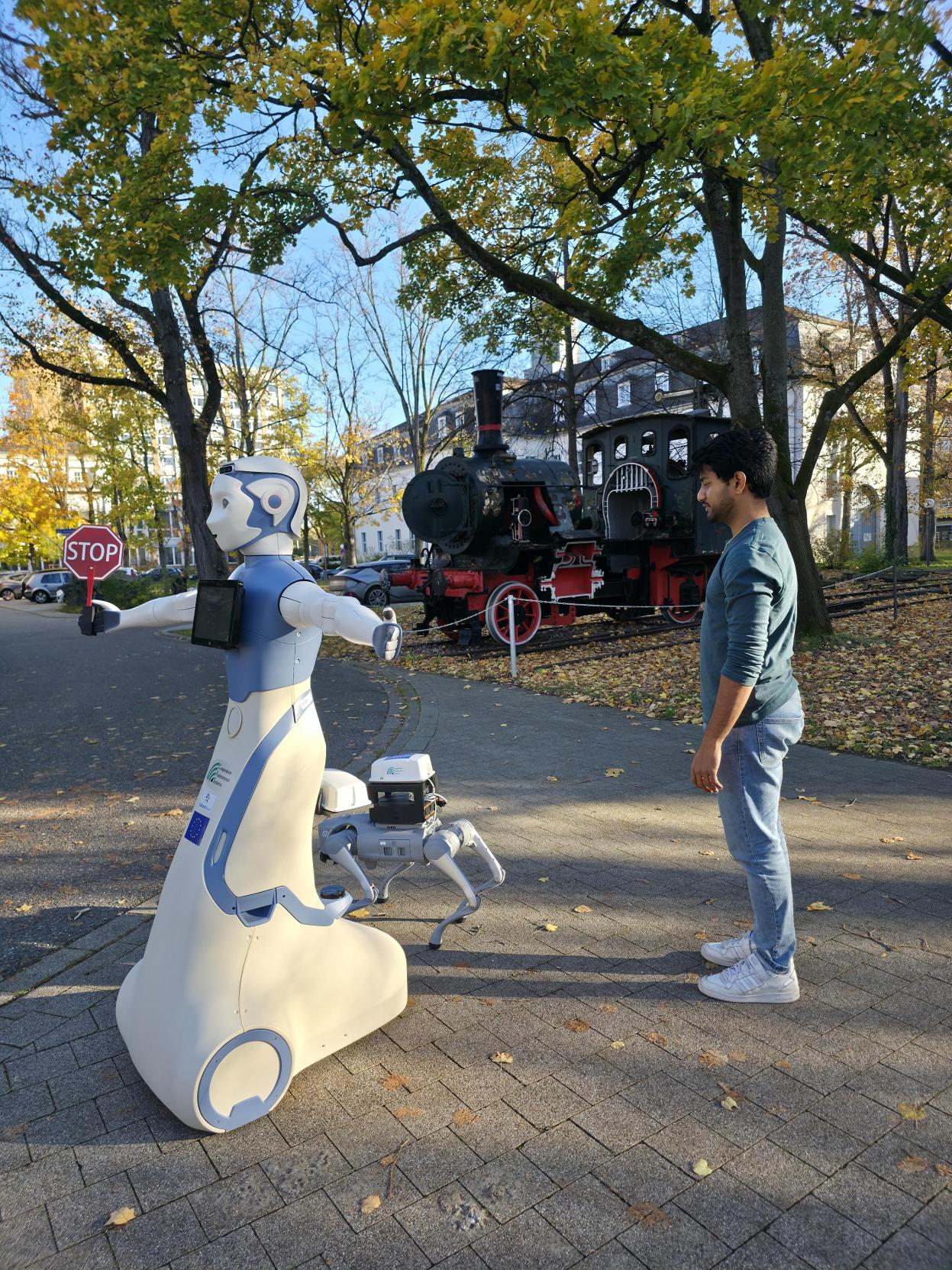}%
         \label{fig:stage3}}
    \hfill
    \subfloat[Stage 4: Maneuver Execution]{%
        \includegraphics[width=0.19\textwidth]{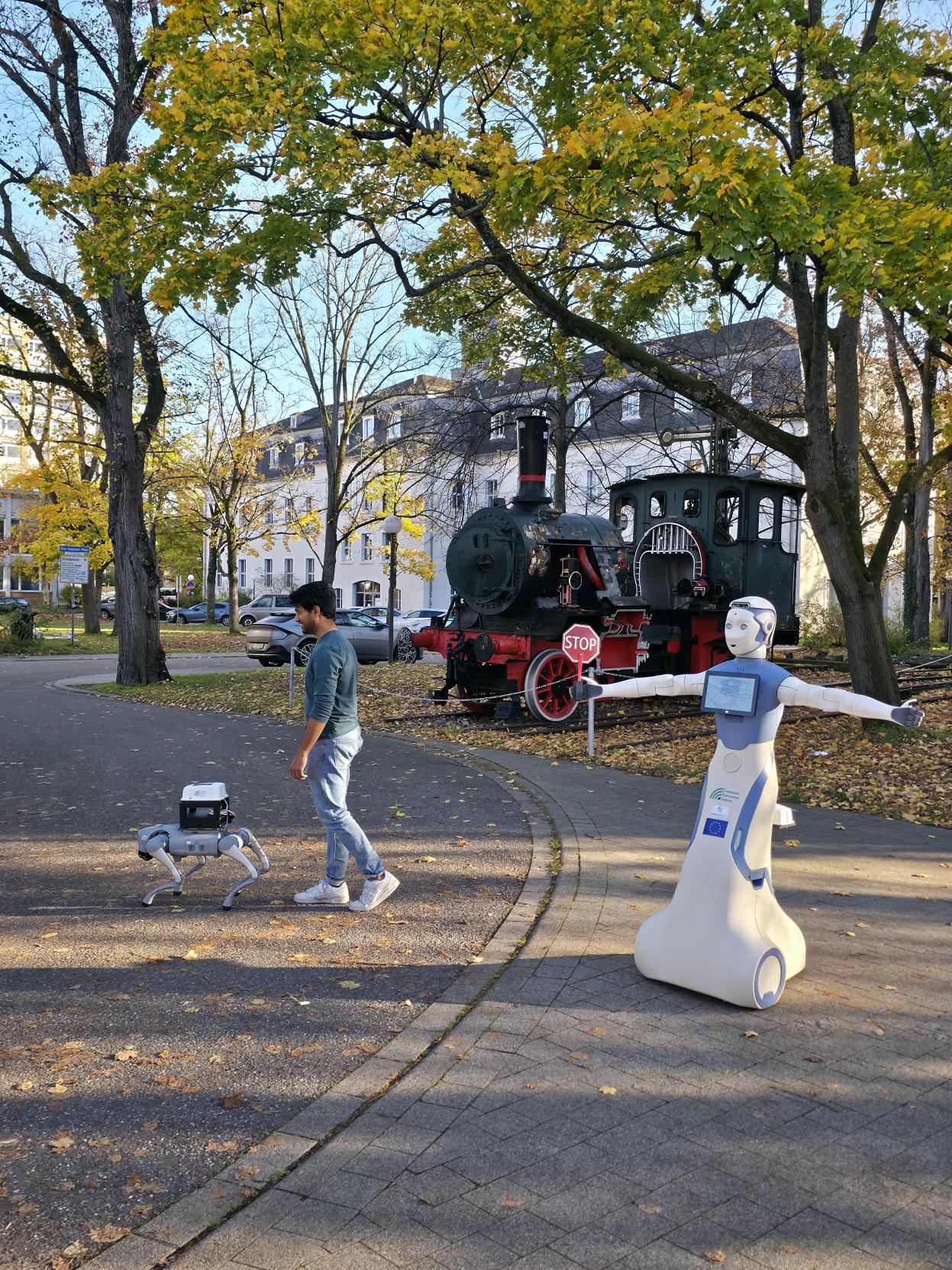}
        \label{fig:stage4}}
    \hfill
    \subfloat[Stage 5: Termination]{%
        \includegraphics[width=0.19\textwidth]{figures/Stage_1.jpg}}

    \caption{\small Five-stage multi-robot coordination using V2X OBUs (white boxes) for pedestrian crossing assistance with RAM and RMCM.}
    \label{fig:coordination_process}
\end{figure*}

Fig.~\ref{fig:coordination_process} illustrates the five-stage coordination sequence realized during the pedestrian assistance task.

\paragraph{Stage 1: Initialization (Idle State)}
Both robots begin in the \textbf{Idle} state, periodically broadcasting RAM that advertise their position, operational status, and capabilities. No coordination relationship is active. This stage reflects idle robot operation as a passive traffic participant and persists until a pedestrian is detected.

\paragraph{Stage 2: Help Establishment (HelpRequested State)}
Upon detecting a pedestrian in a safety-critical crossing zone, ARI issues visual stop gestures and auditory commands, and triggers the \texttt{PedestrianDetected} event and transitions to the \textbf{HelpRequested} state. ARI updates its RAM with \texttt{helpStatus = needHelp}, explicitly requesting for assistance. RoboDog receives this request and responds with a RAM indicating \texttt{helpStatus = offerHelp}, thereby advertising its availability to support the task.

\paragraph{Stage 3: Role Assignment (RoleEstablished State)}
After receiving the assistance offer, ARI confirms the leader-follower relationship via RAM, triggering the \texttt{RoleConfirmed} event. Both robots transition to the \textbf{RoleEstablished} state, with ARI taking the leader role and RoboDog taking the follower role. ARI now instructs the pedestrian to follow RoboDog for guided crossing using auditory cues and turns to manage oncoming traffic with visual stop gesture. At this stage, roles are fixed, but no maneuver is yet active. Periodic RAM updates maintain synchronization and situational awareness.

\paragraph{Stage 4: Maneuver Execution (ManeuverExecuting State)}
While in the \textbf{RoleEstablished} state, ARI issues one or more RMCM commands. Upon receiving an RMCM, RoboDog immediately responds with an RMCM acknowledgment, confirming successful reception of the maneuver instruction. Following this acknowledgment, RoboDog transitions into the \textbf{ManeuverExecuting} state and physically executes the commanded maneuver.

In the POC, when ARI determines it is safe to cross, two maneuvers are executed sequentially. First, an RMCM with \texttt{taskType = initialPos} instructs RoboDog to move into a designated lead position. Second, an RMCM with \texttt{taskType = Move} initiates pedestrian escort across the roadway. Completion of each maneuver is not signaled via RMCM but is inferred by the leader through subsequent RAM from the follower, which reflect updated position, motion state, and task progression. Once \texttt{ManeuverCompleted} is detected via RAM, both robots transition back to the \textbf{RoleEstablished} state, allowing multiple coordinated tasks to be executed under a single role association.

\paragraph{Stage 5: Termination (Return to Idle State)}
Once all required maneuvers are completed meaning that the pedestrian has safely crossed, ARI triggers the \texttt{TaskCompleted} event. The leader-follower association is explicitly released via \texttt{helpStatus = None} in RAM, and both robots transition back to the \textbf{Idle} state, resuming RAM periodic broadcasts and remaining available for future tasks.


To quantify coordination efficiency, we adopt the \emph{total
maneuver coordination time} (TMCT), following the
decomposition into negotiation and execution time introduced
by~\cite{maksimovski2024framework}.
Per trial,
\begin{equation}
  T_\mathrm{TMCT} \;=\; \underbrace{(t_\mathrm{rx}-t_\mathrm{tx})}_{T_\mathrm{neg}}
  \;+\;
  {T_\mathrm{exec}},
  \label{eq:tmct}
\end{equation}
where $t_\mathrm{tx}$ is the leader's RMCM transmission time,
$t_\mathrm{rx}$ the follower's reception time, and $T_\mathrm{exec}$ the elapsed time until the follower executes \texttt{ManeuverCompleted}. Across $N = 5$ trials, \texttt{initialPos} yielded 
$T_\mathrm{neg} = 0.074 \pm 0.015\,\mathrm{s}$ and 
$T_\mathrm{exec} \approx 0.50\,\mathrm{s}$ 
($T_\mathrm{TMCT} = 0.574 \pm 0.015\,\mathrm{s}$); the pedestrian-escort \texttt{Move} maneuver yielded 
$T_\mathrm{neg} = 0.129 \pm 0.102\,\mathrm{s}$ and 
$T_\mathrm{exec} = 21.60\,\mathrm{s}$ 
($T_\mathrm{TMCT} = 21.729 \pm 0.102\,\mathrm{s}$), demonstrating stable
and low-latency coordination timing across trials. 

Reliability is ensured through subsequent RAM updates and explicit RMCM acknowledgments, consistent with ETSI CCAM design principles. Missed RAM is compensated by subsequent broadcasts, while RMCM commands require acknowledgments from the follower. In the absence of command from the leader within a safety timeout, the follower transitions to a safe idle position and advertises its status via RAM, enabling the leader to reissue commands or the follower request to issue leave coordination.

\subsubsection*{Extending FSM Control}
In the presented POC, the five-stage FSM enables decentralized, low-latency, and reliable coordination between two heterogeneous robots assisting a single pedestrian. This baseline scenario highlights a practical limitation: with one humanoid acting as leader, only a single traffic direction can be actively blocked as seen in Fig.~\ref{fig:stage4}. To support bilateral traffic flows, the FSM can be extended to a two-humanoid deployment, wherein the pedestrian detecting humanoid remains the leader and assigns a second humanoid as follower via the \emph{Help Establishment} and \emph{Role Assignment} stages. During the \emph{Maneuver Execution} stage, the leader can issue \texttt{initialPos} and \texttt{block} commands to the follower through RMCM, forming a bidirectional safety corridor that allows pedestrians from both directions to cross while maintaining synchronization through periodic RAM updates.

The FSM can further accommodate a third robot, such as RoboDog, without structural modification. By leveraging the same \emph{Help Establishment} and \emph{Role Assignment} stages, the leader can simultaneously send RMCM maneuvers to multiple followers, assigning one humanoid to traffic blocking and RoboDog to pedestrian escort. This demonstrates that coordination efficiency and reliability are preserved as team size increases. By avoiding centralized control and fixed team assumptions, the event-driven FSM offers a scalable, robust, and flexible basis for multi-robot coordination.


\subsection{VRU Clustering and Scalability Evaluation}
\label{sec:vru_clustering_illustration}

To further evaluate scalability and network efficiency, this section focuses on VRU integration in mixed V2X environments. Specifically, it examines how robots acting as mobile sensing and communication agents can cluster nearby non-V2X VRUs in high-risk intersections, and disseminate V2X-Enabled VRUs collective state using RAM, thereby reducing channel load while maintaining safety-critical awareness.

\subsubsection{Scenario Setup}
The evaluation was conducted using a $3 \times 3$ Manhattan grid representing a low-speed urban environment, where each road segment has a length of 100~m. Sidewalks are present along both sides of the roads, while pedestrian crossings are provided only near intersections to emulate realistic urban behavior. The scenario was implemented using the \textit{SUMO traffic simulator}~\cite{sumo} coupled with the
\textit{Artery}~\cite{riebl2015artery} V2X networking stack. The V2X communication uses the standard ETSI ITS-G5 stack
(IEEE~802.11p/EDCA, 5.9\,GHz, 10\,MHz channel, 200\,mW
transmission power). 


The scenario includes mixed traffic participants consisting of pedestrians and vehicles. Robots equipped with V2X communication capabilities are placed at grid intersections, where pedestrian-vehicle interactions are most frequent and safety-critical, and operate as mobile traffic mediators. Three deployment configurations are considered: \textit{No robot} (baseline), \textit{1 robot} at the central intersection $(2,2)$, and \textit{9 robots} with one robot at each intersection in the grid.

Each robot is assigned a circular observation zone with two evaluated sensing radii: 15~m, representing optimistic outdoor pedestrian detection capabilities~\cite{torres2023pedestrian}, and 10~m, reflecting more conservative sensing assumptions. Simulations were conducted with 20, 50, and 100 pedestrians and a fixed set of 20 vehicles. Pedestrian routes were randomly generated but kept identical across all robot configurations to ensure comparability and reproducibility in mixed-traffic conditions. The evaluation intentionally focuses on intersection-centric mediation with static robot positions and bounded sensing ranges, isolating how robot-mediated clustering enhances awareness and reduces channel load in safety-critical regions under realistic CCAM deployment constraints.

\subsubsection{Case 1: Non-V2X Pedestrians}
\label{sec:case1_non_v2x}

In this case, pedestrians are configured in \textit{Artery} as non-V2X-enabled and therefore do not transmit VAMs. When entering a robot’s observation zone, pedestrians are detected by onboard sensors and incorporated into the V2X ecosystem using the \texttt{VruClusterInformationContainer} in RAM. To quantify the ability of robots to integrate non-V2X pedestrians into V2X, the \emph{Observation coverage ratio (OBS)} is evaluated. This metric measures the fraction of pedestrian travel time during which pedestrians are within a robot’s observation zone, normalized by total pedestrian travel time.



\paragraph{Metric Definition}
Let $T_{\text{obs}}$ denote the cumulative time during which any of those pedestrians are within robot observation zones, and let $T_{\text{sim}}$ denote sum of travel times of all pedestrians in the scenario. The \emph{OBS} is defined as:
\begin{equation}
C_{\text{obs}} = \frac{T_{\text{obs}}}{T_{\text{sim}}} \times 100
\end{equation}

\begin{figure}[t]
\centering
\includegraphics[width=0.8\columnwidth]{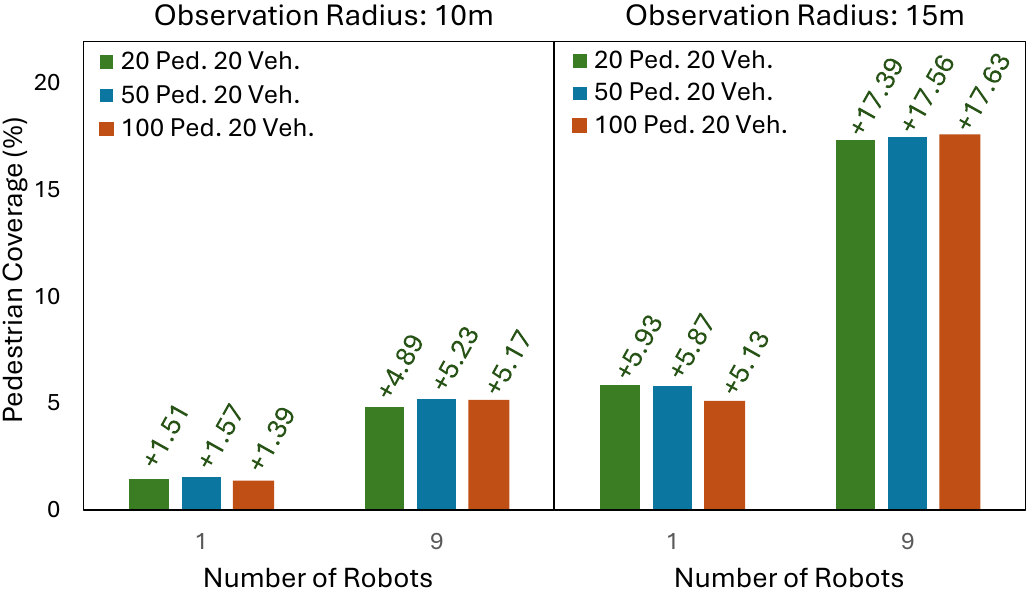}
\caption{Observation coverage ratio (OBS) for non-V2X pedestrians under varying robot deployment densities and observation radii of 10 m and 15 m.}
\label{fig:coverage}
\end{figure}

As is well established, no pedestrian coverage can be achieved in the absence of robots when pedestrians do not generate VAMs; therefore, the \textit{No robot} baseline is omitted from Fig.~\ref{fig:coverage}, as it would yield 0\% coverage across all scenarios. Introducing robots significantly increases VRU coverage, even with limited sensing ranges,  as illustrated in Fig.~\ref{fig:coverage}. With a 15~m observation radius, \textit{1 robot} achieves approximately 5--6\% coverage, while deploying \textit{9 robots} increases coverage to approximately 17--18\%. Importantly, these coverages occur predominantly at intersections, which constitute the most safety-critical segments of pedestrian trajectories.

Reducing the observation radius to 10~m leads to a proportional decrease in coverage across all configurations; however, the relative trends remain consistent. Multi-robot deployment continues to substantially outperform single-robot placement, demonstrating that even conservative sensing ranges are sufficient to extend V2X awareness to non-equipped VRUs in safety-critical regions. These results highlight that robot-mediated awareness need not provide full trajectory coverage, but can strategically target high-risk locations in line with CCAM safety objectives.


\subsubsection{Case 2: V2X-Enabled Pedestrians}
\label{sec:case2_v2x}

In this case, pedestrians and vehicles are configured as V2X-enabled in \textit{Artery} and transmit standard VAMs and CAMs, respectively. Upon entering a robot’s observation zone, pedestrians are incorporated into \texttt{VruClusterInformationContainer} in RAM, with the robot acting as the cluster head. Upon receiving the clustered information disseminated via RAM, pedestrians suspend their individual VAM transmissions in accordance with ETSI VAM clustering principles~\cite{etsi2023ts103300-3}, thereby reducing redundant message generations.

The impact of robot-mediated clustering is evaluated using the \textit{Channel Busy Ratio (CBR)}, defined as the fraction of time the wireless channel is sensed as busy~\cite{etsi2021cbr}. We report the \textit{mean Channel Busy Ratio (mCBR)}, representing the time-averaged CBR over the entire simulation duration.

\paragraph{mCBR Reduction Metric}
The relative reduction in channel load due to robot-enabled clustering is computed as:
\begin{equation}
\text{mCBR Reduction (\%)} =
\frac{\text{mCBR}_{\text{NoRobot}} - \text{mCBR}_{\text{Robot}}}
{\text{mCBR}_{\text{NoRobot}}}
\times 100
\end{equation}





\begin{figure}[t]
\centering
\includegraphics[width=0.85\columnwidth]{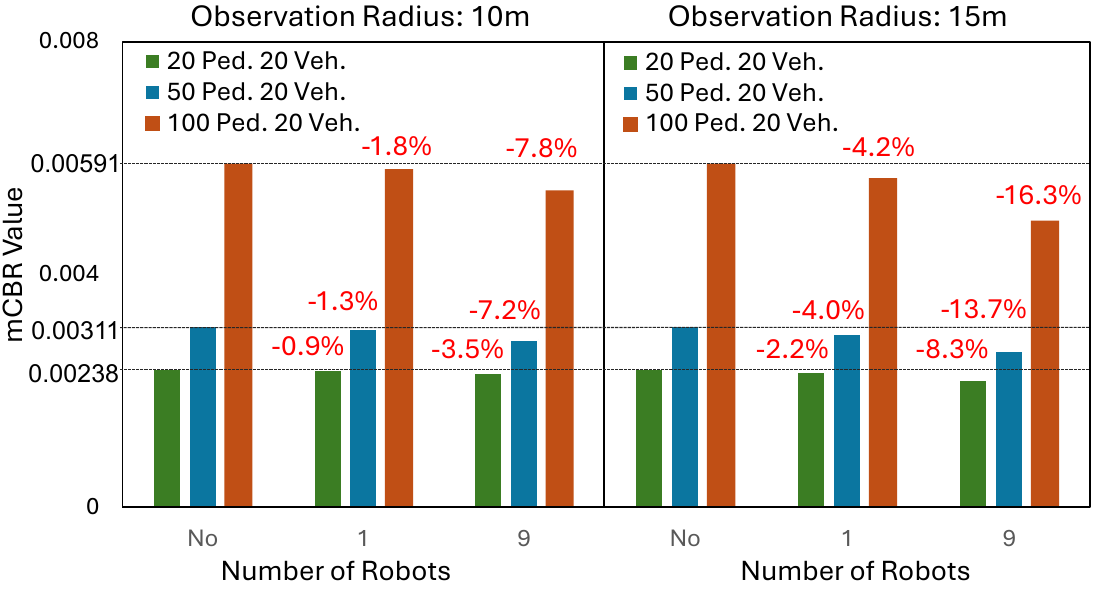}
\caption{Mean channel busy ratio (mCBR) values from robot-mediated VRU clustering versus \textit{No robot} baseline for observation radii of 10 m and 15 m.}
\label{fig:mcbr}
\end{figure}

As illustrated in Fig.~\ref{fig:mcbr}, the measured mCBR remains comparatively low across all evaluated scenarios with vehicles present. Robot-mediated clustering consistently reduces mCBR compared to the \textit{No robot} baseline across all pedestrian densities. For a 15~m observation radius, clustering yields increasingly pronounced mCBR reductions as pedestrian density grows, reaching up to 16.3\% with \textit{9 robots} in the 100-pedestrian scenario. When the observation radius is reduced to 10~m, the absolute reduction decreases; nevertheless, a clear and consistent benefit is preserved across all configurations.




This comparison highlights an important system-level trade-off. Larger observation radii improve clustering effectiveness and channel load reduction by aggregating a larger fraction of the VRU population, while smaller radii limit sensing scope and communication overhead in a mixed traffic environment. Moreover, the diminishing marginal gains observed with increasing robot density indicate that effective VRU clustering does not require dense robot deployments. These findings support cost-efficient and adaptive robot placement strategies that prioritize safety-critical intersections rather than uniform urban coverage.

\section{Conclusion}
This paper presented a V2X-based framework that enables robots to operate as active and cooperative participants within the CCAM ecosystem. By extending ETSI facility-layer services through the RAS and RMCS--realized via RAM and RMCM--the framework supports decentralized multi-robot coordination, explicit role assignment, and the integration of both V2X-enabled and non-V2X VRUs, without reliance on centralized infrastructure.

A real-world POC demonstrated that heterogeneous robots (ARI and RoboDog) can coordinate safely, deterministically, and with low latency to assist a pedestrian during a road-crossing scenario using only V2X communication. In addition, simulation studies showed that robot-mediated VRU clustering improves cooperative awareness in high-risk areas by integrating non-V2X pedestrians while reducing redundant transmissions from V2X-enabled pedestrians, thereby enhancing overall network efficiency. Beyond feasibility, these results provide actionable guidance for ITS practitioners, including indicative robot deployment densities, observation radius trade-offs, and VRU clustering strategies that balance awareness benefits against channel load. These insights support informed design decisions for integrating mobile robots into real-world CCAM deployments.



The framework is grounded in a formally specified, event-driven finite-state coordination model, validated through on-hardware experiments and ETSI-compliant implementations, ensuring predictable behavior. From a practical standpoint, the evaluated scenarios demonstrate direct relevance to real-world VRU safety.

While these results demonstrate technical soundness and practical relevance for scenarios such as school crossings, mixed traffic, and VRU safety, the current evaluation is limited to a finite set of real-world traffic scenarios and controlled simulations with a fixed $3\times3$ grid, static robot positions, no overlap zone, and predefined observation radii. Large-scale deployments, dynamic mobility, and feasibility under large-team studies remain future work. Further work will address robot-to-vehicle maneuver coordination, and broader challenges related to, security, privacy, public acceptance and trust in robot-assisted traffic mediation~\cite{Bied2025}.



\section{Acknowledgment}

This work was funded by the KIT Future Fields Wild Ideas project “WildRobot” and the KIT Future Fields Stage 2 project “V2X4Robot”. Support from the EU-funded “CulturalRoad” project (EU Grant No. 101147397), the Helmholtz programs EDF and ESD, as well as ELLIIT, is gratefully acknowledged.


\bibliographystyle{IEEEtran}
\bibliography{robots}\textbf{}


\end{document}